\def\CleanCopy{1}

\documentclass[pdflatex,sn-nature]{sn-jnl}

\usepackage{graphicx}%
\usepackage{multirow}%
\usepackage{amsmath,amssymb,amsfonts}%
\usepackage{amsthm}%
\usepackage[table]{xcolor}%
\usepackage{textcomp}%
\usepackage{manyfoot}%
\usepackage{booktabs}%
\usepackage{algorithm}%
\usepackage{algorithmicx}%
\usepackage{algpseudocode}%
\usepackage{listings}%
\definecolor{RevisionBlue}{HTML}{285A8E}%
\newcommand{\rev}[1]{{\color{RevisionBlue}#1}}%
\renewcommand{\rev}[1]{#1}
\newcommand{\revtwo}[1]{{\color{RevisionBlue}#1}}%
\ifdefined\CleanCopy\renewcommand{\revtwo}[1]{#1}\fi
\newenvironment{revision}{\color{RevisionBlue}}{}%
\renewenvironment{revision}{}{}%
\theoremstyle{thmstyleone}%
\theoremstyle{thmstyletwo}%

\theoremstyle{thmstylethree}%

\usepackage{float}
\makeatletter
\renewcommand{\addressfont}{\reset@font\fontsize{10bp}{12bp}\selectfont\titraggedcenter}
\makeatother
\begin{document}

\title{Vision-language models lag human performance on physical dynamics and intent reasoning}


\author[1]{\fnm{Tianjun} \sur{Gu}}

\author[1]{\fnm{Jingyu} \sur{Gong}}

\author[1]{\fnm{Zhizhong} \sur{Zhang}}

\author[1,2]{\fnm{Yuan} \sur{Xie}}

\author[1]{\fnm{Lizhuang} \sur{Ma}}

\author*[1]{\fnm{Xin} \sur{Tan}}\email{xtan@cs.ecnu.edu.cn}

\author[3,4]{\fnm{Athanasios V.} \sur{Vasilakos}}

\affil*[1]{\orgname{East China Normal University}, \orgaddress{\city{Shanghai}, \country{China}}}

\affil[2]{\orgname{Shanghai Innovation Institute}, \orgaddress{\city{Shanghai}, \country{China}}}

\affil[3]{\orgname{Computer Science Center (National Supercomputer Center in Jinan), Qilu University of Technology (Shandong Academy of Sciences)}, \orgaddress{\city{Jinan}, \country{China}}}

\affil[4]{\orgname{Shandong Provincial Key Laboratory of Computing Power Internet and Service Computing, Shandong Fundamental Research Center for Computer Science}, \orgaddress{\city{Jinan}, \country{China}}}


\abstract{Spatial intelligence is central to embodied cognition, yet contemporary AI systems still struggle to reason about physical interactions in open-world human environments. Despite strong performance on controlled benchmarks, vision-language models often fail to jointly model physical dynamics, reference frames, and the latent human intentions that drive spatial change. \revtwo{We introduce Teleo-Spatial Intelligence (TSI), which captures an important aspect of spatial reasoning: integrating physical dynamics and action goals to reason about actions and their spatial consequences.} To evaluate TSI, we present EscherVerse, a large-scale open-world resource built from 11,328 real-world videos, including an 8,000-example benchmark and a 35,963-example instruction-tuning set. Across 27 state-of-the-art vision-language models and an independent analysis of first-pass human responses from 11 annotators, we identify a persistent teleo-spatial reasoning gap: the strongest proprietary model achieves 57.26\% overall accuracy, far below first-pass human performance, which ranges from 84.81\% to 95.14\% with a mean of 90.62\%. Fine-tuning on real-world, intent-aware data narrows this gap for open-weight models, but does not close it. EscherVerse provides a diagnostic testbed for purpose-aware spatial reasoning and highlights a critical gap between pattern recognition and human-level understanding in embodied AI.}

\maketitle

\section*{Introduction}

{\sloppy\looseness=-1
In recent years, Artificial Intelligence (AI) has demonstrated formidable capabilities across multiple domains. Represented by GPT~\cite{brown2020language, openai2023gpt4} and AlphaFold~\cite{jumper2021highly}, AI has achieved human-expert levels, and even surpassed them in tackling complex problems. This progress has significantly bolstered public trust in AI and accelerated its deployment into the real world, where it participates in complex decision-making and physical interactions~\cite{brohan2022rt1, brohan2023rt2}. However, when AI systems are deployed in physical environments shared with humans, their outputs often directly or indirectly influence safety-critical spatial decisions. Such decisions rely heavily on spatial intelligence~\cite{hegarty2010components}, the ability to comprehend spatial changes within physical constraints and anticipate human intent. By contrast, although current AI demonstrates formidable capabilities in language and visual tasks, its understanding of spatial dynamics often remains limited. It still struggles to reason about dynamic environments and human intent, posing potential risks in real-world deployment.
\par}

According to a recent analysis of the U.S. Occupational Safety and Health Administration (OSHA) severe injury reports, 77 robot-related accidents occurred between 2015 and 2022, while a National Institute for Occupational Safety and Health (NIOSH) study identified 41 robot-related fatalities in the physical workplace between 1992 and 2017~\cite{layne2023robot, sanders2024robot}. In the domain of autonomous driving, the National Highway Traffic Safety Administration (NHTSA) reported that between July 2021 and May 2022 alone, vehicles equipped with automated driving and driver assistance systems were involved in over 500 reported crashes, resulting in multiple fatalities and serious injuries, including collisions with vulnerable road users~\cite{nhtsa2022sgo,nhtsa2022adas}. \rev{These reports underscore the practical importance of reliable perception and decision-making in physical environments. Although they do not isolate a single computational cause, they motivate systems that combine reliable perception with reasoning about spatial change and human behaviour in complex, dynamic environments.}

Existing research typically defines spatial intelligence as the ability to reason about the location, geometric relationships, or motion trajectories of objects~\cite{Johnson_2017_CVPR,yang2019spatialsense}. This definition largely overlooks the intentional factors driving spatial change—namely, why humans take specific actions and how these actions influence future spatial states. From the perspective of cognitive science, human spatial cognition goes fundamentally beyond passive geometric perception; it is deeply rooted in teleological reasoning and Theory of Mind (ToM) \cite{baker2009action, spelke2007core}. Humans instinctively parse dynamic scenes not merely as kinematic transformations, but as sequences of intent-driven behaviors constrained by physical environments and spatial affordances \cite{gibson2014ecological}. We contend that this understanding of spatial intelligence remains incomplete. To achieve predictable and safe behavior in the physical world, AI must simultaneously comprehend physical dynamics and the latent human intentions driving spatial change. 

\revtwo{Therefore, we introduce Teleo-Spatial Intelligence (TSI), derived from the Greek \textit{telos} (goal), to integrate physical dynamics and goal-directed spatial reasoning in studying how human goals shape actions and their spatial consequences.} TSI does not replace existing spatial reasoning frameworks but emphasizes \rev{two complementary and often interacting components}: 1) understanding object dynamics and interactions within physical constraints; and 2) inferring the human intentions and goals driving these spatial changes. This perspective expands spatial reasoning from what happened to why it happened, aligning more closely with human decision-making in real-world environments.  

\rev{For analysis, we treat these components as related but separately measurable rather than as one indivisible ability. We distinguish physical-dynamic content, dependence on observable goals and questions requiring both.}

\begin{figure}[tbp]
\centering
\includegraphics[width=\textwidth,height=.70\textheight,keepaspectratio]{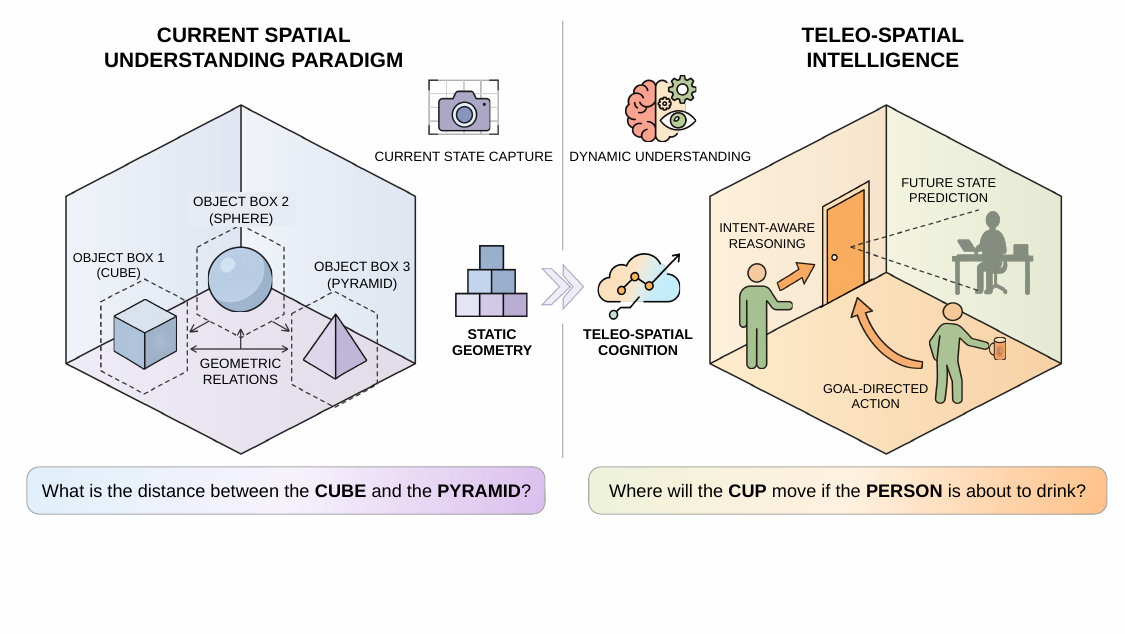}
\caption{\textbf{Teleo-Spatial Intelligence (TSI) contrasted with the current spatial understanding paradigm.} Left, conventional spatial understanding is primarily object-centred, geometry-focused, and grounded in current-state description. Right, TSI extends spatial reasoning toward human-centred, goal-directed, and intent-aware understanding of future spatial change. Example questions illustrate the difference between static geometric queries and intent-conditioned reasoning about dynamic outcomes.}
\label{fig:tsi_concept}
\end{figure}

{\sloppy\looseness=-2
Figure~\ref{fig:tsi_concept} contrasts this notion with the dominant object-centred spatial paradigm and illustrates how teleo-spatial reasoning links current observations to goal-directed future change.
\par}

Widely used visual question-answering benchmarks~\cite{goyal2017vqa,hudson2019gqa,liu2023mmbench} provide broad measures of visual understanding, but do not specifically isolate how physical dynamics and observable human goals interact in changing scenes. To address this gap, we introduce EscherVerse, a large-scale open-world benchmark for Teleo-Spatial Intelligence. By designing problems grounded in dynamic scenes and human intent, this benchmark systematically evaluates models' capabilities in both physical-dynamics comprehension and intent inference, helping clarify the capability boundaries of artificial intelligence in real spatial environments.

\rev{Recent Theory-of-Mind benchmarks, including ToMBench, MMToM-QA, EgoToM and MuMA-ToM, examine beliefs, goals and plans in text or multimodal activity~\cite{chen2024tombench,jin2024mmtom,li2025egotom,shi2025muma}. Spatial benchmarks such as SpatialVLM, VSI-Bench and Spatial-DISE instead emphasize metric relations, three-dimensional structure, motion and reference frames~\cite{chen2024spatialvlm,yang2025vsibench,huang2025spatialdise}. EscherVerse complements both lines by testing open-world video questions in which tracking, state change, viewpoint transformation and prediction can depend on an observable goal. Supplementary Table~1 provides a direct comparison.}

Recent work has extended evaluation to long-horizon world-state tracking in real videos (VSI-Super-Wild)~\cite{gu2026supersensing}, multi-hop spatial deduction (Video-MSR)~\cite{zhu2026videomsr}, and spatial relations combined with object affordances and context-dependent utility (SFI-Bench)~\cite{zhang2026sfi}. These complementary directions broaden the scope of spatial evaluation. EscherVerse focuses on how observable action goals interact with physical dynamics, tracking and viewpoint changes in human-centred video scenes.

We make three main contributions. First, we introduce Teleo-Spatial Intelligence (TSI) as a perspective on reasoning about physical dynamics, reference-frame transformations, and goal-directed human action in open-world scenes, extending conventional geometry-centred views of spatial intelligence. Second, we operationalize this perspective in EscherVerse, a large-scale real-world resource built from 11,328 videos, including an 8,000-example benchmark and a 35,963-example instruction-tuning set, enabling systematic evaluation and training for teleo-spatial reasoning. Third, through a systematic study of 27 frontier vision-language models and independent first-pass human responses from 11 annotators, we establish a large and persistent human-model gap on these tasks, and show that real-world, intent-aware supervision substantially improves open-weight models while still leaving performance far below human level.

\section*{Results}

\subsection*{Overview of EscherVerse and key findings}\label{sec2}

\begin{figure}[tbp]
\centering
\includegraphics[width=\textwidth,height=.70\textheight,keepaspectratio]{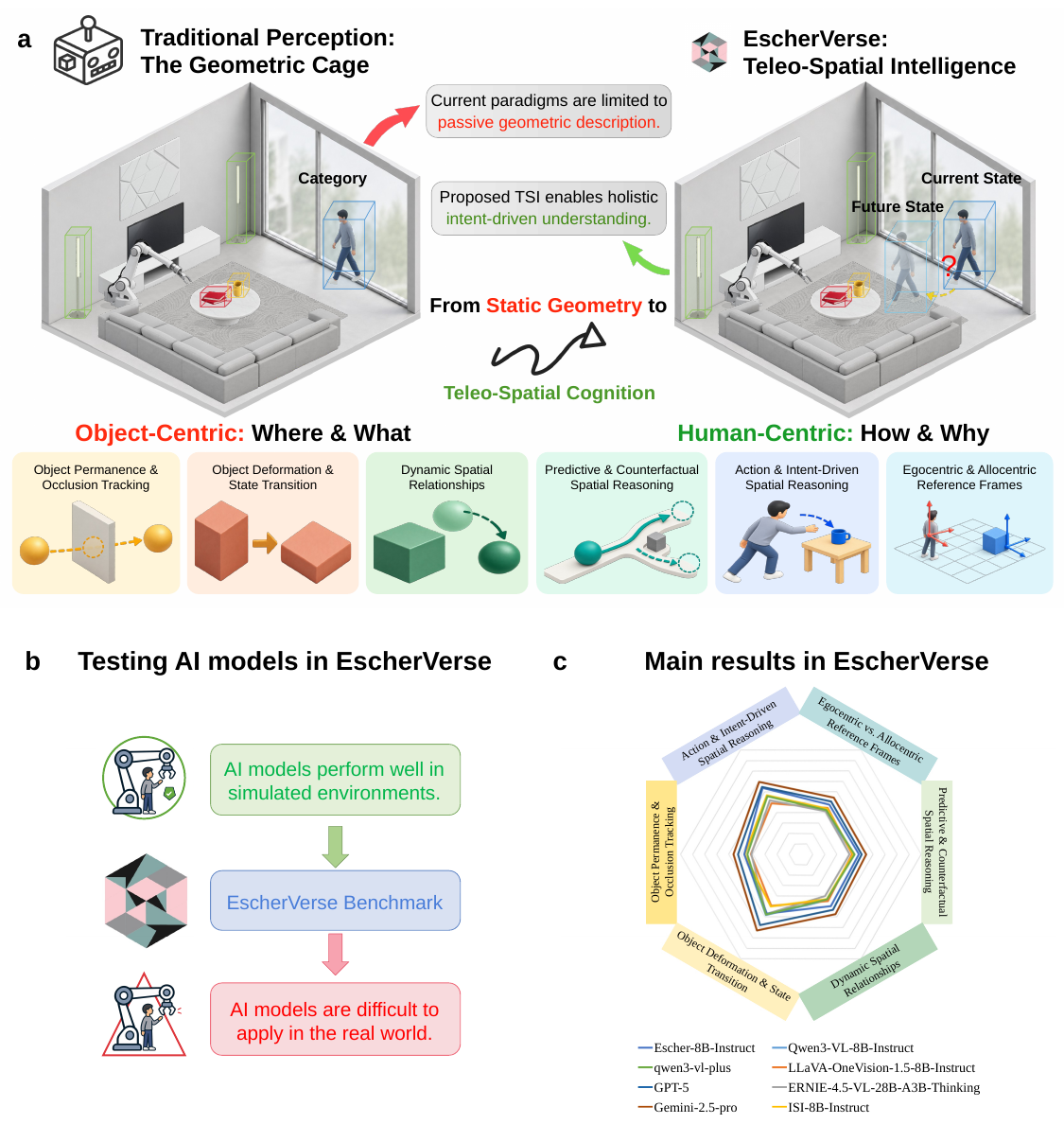}
\caption{Conceptual design and evaluation of EscherVerse. a, Comparison of geometry-centred perception with goal-directed spatial reasoning, with the six benchmark categories shown below. b, Schematic motivation for evaluating models on real-world video tasks. c, Category-level accuracy profiles for the eight models named in the key; each coloured line represents one model. TSI, Teleo-Spatial Intelligence. Source data are provided as a Source Data file.}
\label{fig:fig1}
\end{figure}

\begin{figure}[tbp]
\centering
\includegraphics[width=\textwidth,height=.70\textheight,keepaspectratio]{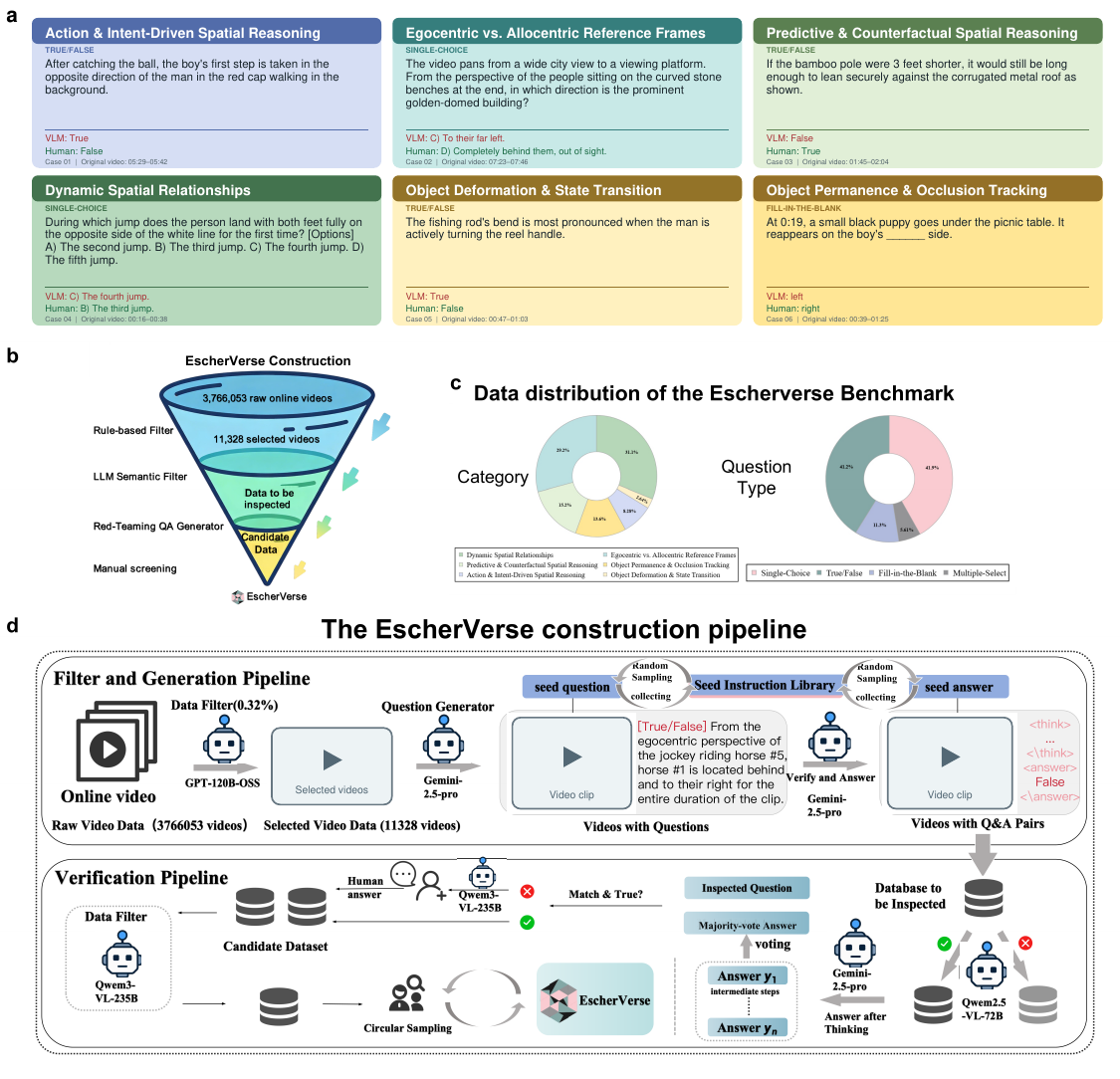}
\caption{EscherVerse examples and construction. a, Example questions from the six categories, presented as text case cards with model and human responses. b, Stages of video selection and question screening. c, Benchmark composition by category (left) and question format (right); coloured sectors identify the categories and formats listed in the keys. d, Question generation and verification pipeline, including model screening and human review; video symbols are schematic. Original video links and clip time ranges for panel a are listed in Supplementary Table 17. QA, question--answer; LLM, large language model; VLM, vision-language model. Source data are provided as a Source Data file.}
\label{fig:dataset}
\end{figure}

EscherVerse assesses Teleo-Spatial Intelligence (TSI) through a diverse and cognitively demanding set of tasks that jointly probe physical dynamics, viewpoint transformations, and human intent in real-world scenes. We curated 11,328 real-world video clips from authentic data sources and organized them into two complementary resources: an 8,000-example benchmark for evaluation and a 35,963-example instruction-tuning set for model development. Together, these data cover six key dimensions of spatial reasoning, from foundational object constancy to advanced intention-driven causality. Figure~\ref{fig:fig1} summarizes the conceptual design, six task categories and evaluation. Representative examples, video selection, benchmark composition and the curation pipeline are shown in Figure~\ref{fig:dataset}a--d, respectively.

In this study, we conducted a systematic assessment of 27 state-of-the-art models, including frontier proprietary systems and diverse open-weight VLMs, to measure the current frontier of TSI. In parallel, we analyzed independent first-pass human responses from 11 annotators collected during benchmark construction before consensus formation. This comparison reveals a clear human-model gap: the strongest proprietary model reaches only 57.26\% overall accuracy, whereas first-pass human performance ranges from 84.81\% to 95.14\% with a mean of 90.62\%. These deficits are accompanied by recurring failure modes, including perspective locking, loss of spatiotemporal continuity, and incorrect binding between actions and goals.

We further show that supervised fine-tuning on the EscherVerse instruction-tuning set yields substantial gains for open-weight models, narrowing but not closing the gap to human performance.

\definecolor{LightGreen}{HTML}{C8E6C9}
\definecolor{MidGreen}{HTML}{A5D6A7}
\definecolor{LightCyan}{HTML}{E0F7FA}
\definecolor{MidCyan}{HTML}{B2EBF2}
\definecolor{LightGray}{gray}{0.92}
\definecolor{MidGray}{gray}{0.88}
\definecolor{DarkGray}{gray}{0.8}

\subsubsection*{Current vision-language models lag independent human performance.}
To systematically evaluate Teleo-Spatial Intelligence (TSI), we conducted a comprehensive assessment across 27 vision-language models, including leading proprietary systems such as GPT-5\cite{singh2025openai}, Gemini-2.5-Pro\cite{comanici2025gemini}, Gemini-2.5-flash and Qwen3-VL-Plus\cite{bai2025qwen3vl}, as well as open-weight families including Qwen-VL\cite{bai2023qwenvl}, LLaVA-NeXT-Video-7B\cite{zhang2024llavanextvideo}, LLaVA-OneVision-1.5\cite{an2025llava}, MiniCPM-V-4\cite{yao2024minicpm}, ERNIE-4.5-VL-28B\cite{ernie2025technicalreport}, Spatial-MLLM\cite{wu2025spatialmllm}, and ViCA2-7B\cite{feng2025towards}. In parallel, we extracted independent first-pass human responses retained during benchmark construction and evaluated human and model performance on the same benchmark items.

{\sloppy\looseness=-1
Across the full 8,000-example benchmark, the 11 independent first-pass annotators achieved accuracies ranging from 84.81\% to 95.14\%, with a mean of 90.62\%. These scores are substantially higher than the strongest proprietary model, Gemini-2.5-Pro, at 57.26\%, and even farther above the strongest open-weight baseline, Qwen3-VL-32B-Thinking, at 49.58\% (Table~\ref{tab:spatial_reasoning_benchmark}). Thus, EscherVerse is challenging but not arbitrarily ambiguous: under a shared annotation protocol, humans reach markedly more reliable judgments than current vision-language models on the same tasks, indicating that current systems remain well below stable human performance on teleo-spatial reasoning.
\par}

\subsubsection*{Data efficiency and capability profiling.}
\begin{figure}[tbp]
\centering
\includegraphics[width=\textwidth,height=.70\textheight,keepaspectratio]{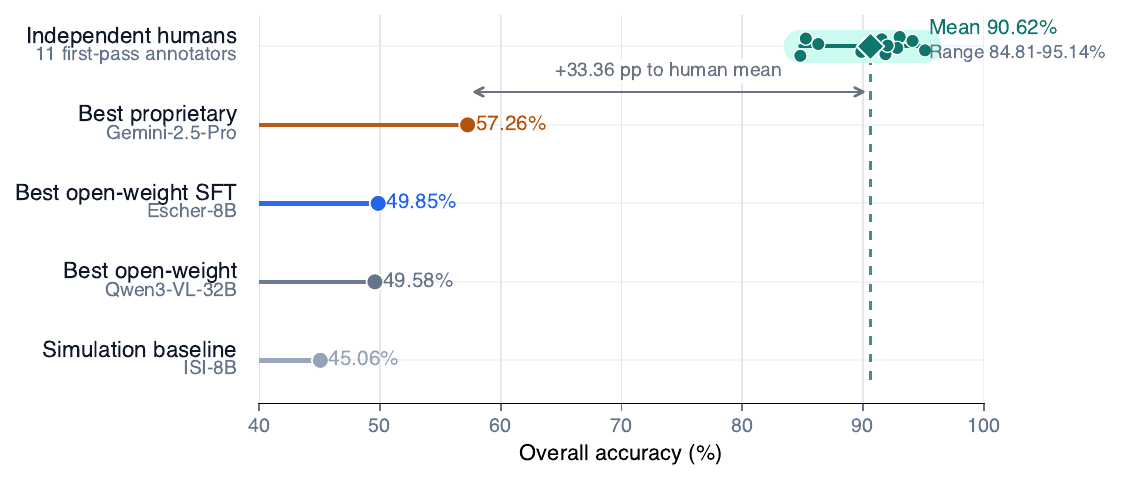}
\caption{\textbf{Independent first-pass human performance exceeds current multimodal models on EscherVerse.} Human performance is shown as the distribution of overall accuracies across 11 annotators, each of whom answered all 8,000 benchmark questions under the first-pass protocol. Current frontier proprietary and open-weight models remain substantially below the human range, with the best proprietary model trailing the human mean by 33.36 percentage points. Source data are provided as a Source Data file.}
\label{fig:human_gap}
\end{figure}

\begin{figure}[tbp]
\centering
\includegraphics[width=\textwidth,height=.70\textheight,keepaspectratio]{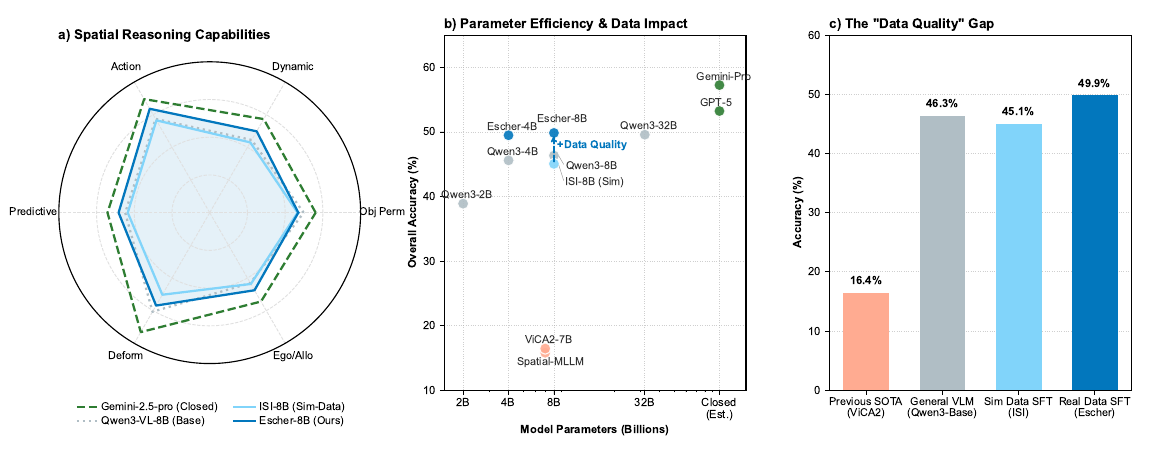}
\caption{The impact of real-world spatial instruction tuning. a, Accuracy profiles across six spatial-reasoning categories for Escher-8B, the simulation-trained ISI-8B baseline, Qwen3-VL-8B and Gemini-2.5-pro; the differences vary by category. b, Overall accuracy plotted against model parameter count for open-weight models. Closed-source models are shown in a separate display category labelled Closed (Est.); their horizontal positions are illustrative and do not represent verified parameter counts. c, Overall accuracy on EscherVerse for ViCA2, the Qwen3 base model, simulation-trained ISI-8B and Escher-8B after real-world instruction tuning. Source data are provided as a Source Data file.}
\label{fig:nature1}
\end{figure}

To evaluate whether this human-model gap can be reduced through better data, we first situate frontier models relative to independent human performance. Figure~\ref{fig:human_gap} shows that the mean first-pass human score reaches 90.62\%, leaving a 33.36 percentage-point gap to the strongest proprietary model. Within the model space, parameter efficiency analysis in Figure~\ref{fig:nature1} b shows that Escher-8B outperforms the significantly larger Qwen3-VL-32B, demonstrating a superior Pareto frontier. As shown in Figure~\ref{fig:nature1} c, our approach establishes a new open-source state-of-the-art (49.9\%), significantly outperforming previous domain-specific models like ViCA2 (16.4\%) and the simulation-based baseline (45.1\%). These gains narrow the gap to human performance, but do not eliminate it.

\subsubsection*{Granular analysis of spatial dynamics.}
{\sloppy\looseness=-2
\rev{We further dissected model performance to examine where the real-world-trained Escher-8B differs from the simulation-trained baseline. The heatmap in Figure~\ref{fig:nature2}a shows larger Escher-8B gains in several dynamics-related categories than in basic recognition.}
\par}

{\sloppy\looseness=-2
The pairwise comparison in Figure~\ref{fig:nature2} b highlights that the switch from simulation (ISI) to real-world data (Escher) yields the largest improvements in Action \& Intent (+7.1\%) and Dynamic Spatial Relationships (+6.8\%), two categories central to teleo-spatial reasoning. Furthermore, robustness analysis in Figure~\ref{fig:nature2} c indicates that our model maintains higher accuracy in challenging multiple-select scenarios, whereas baselines degrade significantly.

\begin{figure}[tbp]
\centering
\includegraphics[width=\textwidth,height=.70\textheight,keepaspectratio]{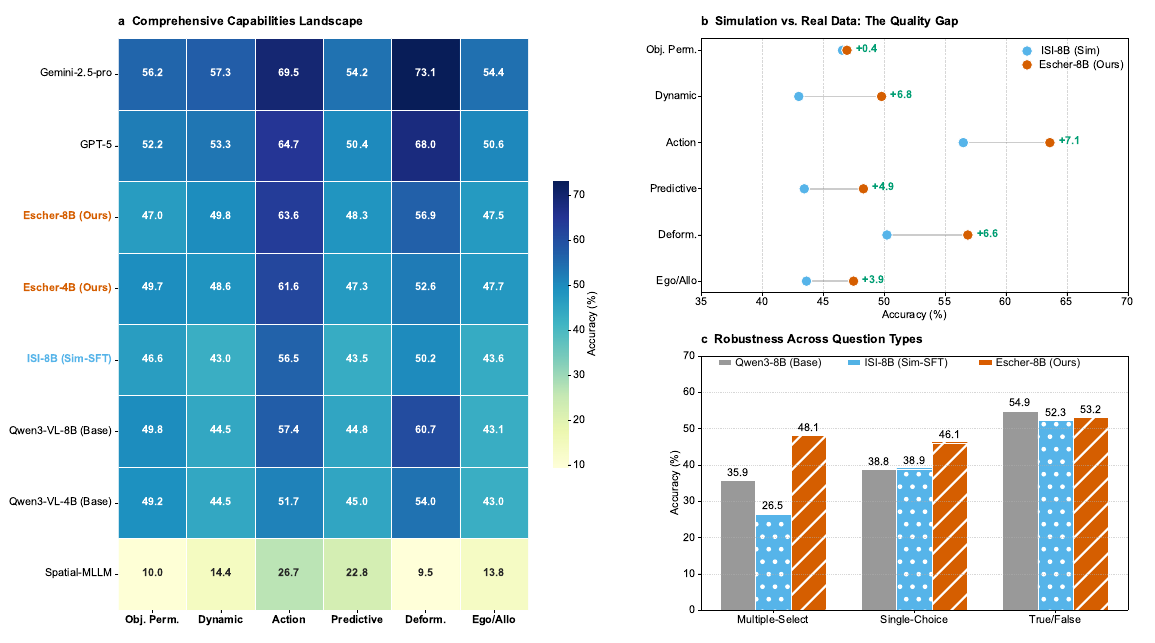}
\caption{\rev{Benchmarking spatial intelligence across diverse modalities and reasoning types. a, Heatmap visualization of accuracy across six spatial-reasoning categories. Escher-8B (orange text) outperforms the simulation-trained ISI-8B (blue text) and reduces the observed gap to the closed-source models in the top rows. b, Category-wise comparison between ISI-8B and Escher-8B. The largest annotated gains are in Action \& Intent and Dynamic Spatial Relationships, followed by Object Deformation, consistent with a benefit from the EscherVerse-35k training data on these benchmark categories. c, Performance by question format. Patterned bars (hatched for Escher and dotted for ISI) show the observed accuracies, including multiple-select questions.} Source data are provided as a Source Data file.}
\label{fig:nature2}
\end{figure}

\par}

\subsubsection*{Behavioral clustering and cognitive failure modes.}
To compare aggregate capability profiles, we performed hierarchical clustering of model scores (Figure~\ref{fig:failures} a). \rev{The two Escher-8B variants first cluster with each other and then with Qwen3-VL-32B-Thinking; this group joins GPT-5 and Gemini-2.5-pro before joining the branch containing ISI-8B. These relationships describe similarity in aggregate performance profiles and do not imply a shared internal reasoning mechanism.}
{\sloppy\looseness=-2
The robustness-complexity landscape in Figure~\ref{fig:failures} b places Escher-8B in the high-robustness regime alongside GPT-5. \rev{Despite these advances, the qualitative cases in Figure~\ref{fig:failures} c, d, e illustrate recurring behavioural errors in tracking simultaneous motions, anchoring dynamic actions and adopting alternative viewpoints.

\begin{figure}[tbp]
\centering
\includegraphics[width=\textwidth,height=.57\textheight,keepaspectratio]{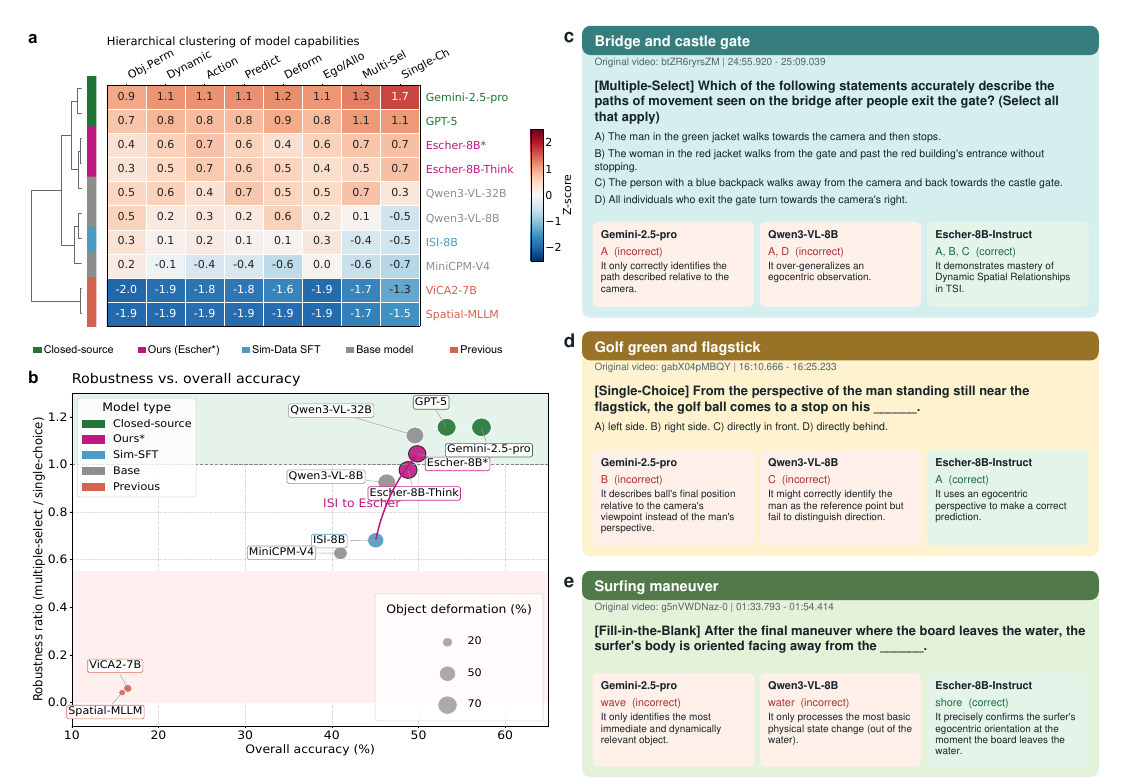}
\caption{\rev{\textbf{Analysis of model behaviours and failure cases.} a, Hierarchical clustering of Table~\ref{tab:spatial_reasoning_benchmark} accuracies across six spatial-reasoning categories and two question formats (multiple-select and single-choice). Each column is standardized across the ten displayed models by subtracting its mean and dividing by its population standard deviation. Ward linkage with Euclidean distance groups the standardized profiles; positive Z-scores indicate above-average accuracy within this model subset. The two Escher-8B variants cluster together, then with Qwen3-VL-32B-Thinking, and subsequently with GPT-5 and Gemini-2.5-pro. These similarities do not establish a shared internal reasoning mechanism. b, The robustness-complexity landscape. The x-axis represents overall accuracy, while the y-axis denotes the Robustness Ratio (accuracy on multiple-select questions divided by accuracy on single-choice questions). Bubble area is proportional to accuracy on Object Deformation. Both panels use the reported aggregate accuracies; the asterisk denotes Escher-8B-Instruct, and Think denotes its Thinking variant. The background zones indicate higher and lower observed robustness ratios. The arrow shows the performance difference between ISI-8B and Escher-8B under these reported metrics. c, Tracking failure: the model does not correctly integrate simultaneous egocentric and allocentric motion. d, Perspective failure: baseline models misidentify the golf ball's final position relative to the person specified in the question. e, Reference-frame failure: baseline models identify the wave or water rather than the shore when asked about the surfer's orientation. Original video links and clip time ranges for panels c--e are listed in Supplementary Table 17.} Source data are provided as a Source Data file.}
\label{fig:failures}
\end{figure}
}
\par}

\begin{revision}
\subsubsection*{Structure of the human--model gap.}
Across the 27 models, all 15 pairwise correlations between category scores were positive (Spearman $\rho=0.753$--$0.961$; median, 0.854; Supplementary Table~2). Principal-component analysis retained one component above the parallel-analysis threshold, accounting for 96.56\% of the variance; Supplementary Table~3 reports the full loadings and leave-one-model-out stability. This common gradient does not make the six content categories interchangeable, so we continue to report them separately.

{\looseness=-2
We next paired 200 benchmark questions with controls testing a necessary visual prerequisite in the same sampled frames. Six of seven models answered these controls with 70.5--86.0\% accuracy, exceeding their accuracy on the corresponding benchmark questions by 32.5--50.5 percentage points (Supplementary Tables~5 and 6; paired McNemar tests, all $P<10^{-10}$). Qwen3-VL-2B was the exception: control and benchmark accuracy were similar (40.5\% and 38.0\%; $P=0.682$). Thus, perception remains limiting for the smallest model, whereas recognizing a necessary visual element is often insufficient for stronger models and motivates analysis of their residual errors.\par}

In a separate same-video analysis, we compared 50 paired questions spanning dynamics versus intent, camera-centred versus actor-goal reasoning, and tracking versus prediction. Three models evaluated after the panel was fixed differed by only 4--8 percentage points between paired sides, and all video-clustered intervals included zero (Supplementary Table~4). This result does not establish a fixed difficulty ordering.

\subsubsection*{Controlled training ablation.}
To test whether the category gains in the original analysis depend on supervision content, we trained Qwen3-VL-4B-Instruct on equal-sized Intent-focused, Dynamic-only and Random-mixture subsets. Each contained 1,409 examples and was matched on question format and answer distribution. Intent-focused training improved the Intent and Dynamic tests by 7.30 and 1.44 points, whereas Dynamic-only training improved them by 4.32 and 5.97 points (Table~\ref{tab:sft_ablation} and Supplementary Tables~11 and 12). The difference-in-differences was 7.51 points (video-clustered bootstrap interval, 3.10--12.17; $P=0.0018$), indicating that supervision content affects where improvement occurs. This result is limited to the fixed 4B/LoRA setting and two training seeds.

\subsubsection*{Quantitative error and scope analyses.}
We classified 300 errors from three models using a predefined codebook and two independent blinded raters, with disagreements adjudicated by a third rater. Primary-label agreement was 79.3\% ($\kappa=0.745$). After adjustment for stratified sampling, the most frequent categories were spatiotemporal grounding (28.8\%) and perspective or reference-frame errors (23.7\%), followed by physical prediction (14.2\%) and perception or recognition (12.1\%; Table~\ref{tab:error_distribution} and Supplementary Table~7). These labels describe observable response patterns rather than internal mechanisms.

\revtwo{In the earlier taxonomy-scope audit, two independent model-based authoring runs proposed questions for 120 videos without access to EscherVerse questions or their category assignments. After evidence checks, 666 question records remained.} Both scope raters judged 516 questions (77.5\%; video-clustered bootstrap interval, 72.8--82.1\%) as directly covered or supported by an operation within the six dimensions; 55 questions (8.3\%) were outside the taxonomy for both raters (Supplementary Tables~8 and 9). \revtwo{This audit measures alignment with the taxonomy; question-sampling coverage instead requires comparison with the questions actually retained for each video.}

\subsubsection*{Task coverage in the retained question bank.}
\revtwo{To examine task sampling within the six dimensions, we compared 200 new human-authored reference questions with 200 original questions and the full 8,000-record bank, using content--operation combinations and subtypes. Under the reported coding, the full bank contains 39 of the 45 fine-grained reference types (86.7\%), compared with 27 in the 200-question original sample. Collapsing subtypes gives 38 of 41 combinations (92.7\%); all 14 reference content categories and all nine reference operations occur in the bank (Supplementary Note~8 and Supplementary Table~13). At an equal sample size of 200 questions, the original sample has greater observed task-type diversity than the human-authored set (74 versus 45 types, with 27 shared; Supplementary Table~14). This advantage is specific to the sampled task types; differences in video composition and question format preclude attributing it solely to the authoring process.}

\revtwo{Five of the six combinations not identified concern evidence sufficiency for contact, crossing, orientation, speed and camera position. The sixth concerns causal explanations of visibility changes with an unspecified mechanism; the bank does contain explanations involving occlusion. Excluding 193 flagged original records leaves the result unchanged at 39/45. Excluding flagged questions from both sets yields 35/41 remaining fine-grained types (85.4\%; Supplementary Tables~15 and 16). These are descriptive comparisons against the sampled reference, not estimates of coverage of every possible task. In the same-clip comparison, 28 of 186 unflagged reference questions (15.1\%) have all their types represented in the original questions for that clip, and 34 (18.3\%) have at least one. Bank-level representation therefore does not imply complete task sampling within individual clips. Supplementary Figures~1--7 show the distributions of all 191 combinations observed across the sets.}

Because hierarchical clustering uses aggregate performance profiles, proximity in Figure~\ref{fig:failures}a should not be interpreted as evidence that models share the same internal reasoning mechanism. The quantitative error analysis above complements the illustrative cases in Figure~\ref{fig:failures}c--e.
\end{revision}





\section*{Discussion}
\rev{Our results indicate that the evaluated vision-language models remain substantially below human performance on teleo-spatial reasoning in open-world scenes. Although Gemini-2.5-Pro is the strongest model in our evaluation, its overall accuracy of 57.26\% remains far below independent first-pass human performance, which ranges from 84.81\% to 95.14\% across 11 annotators. The perception controls show that basic recognition contributes to errors, particularly for the smallest model, but does not account for the full performance gap. The error audit further identifies frequent observable failures in spatiotemporal grounding, reference-frame reasoning, physical prediction, and action--goal binding.}

\rev{The results also clarify the effects of supervision within the evaluated comparisons. Escher-8B outperforms the simulation-trained ISI-8B across the reported aggregate metrics, and the equal-budget ablation shows distinct gains from intent-focused and dynamics-focused subsets. Escher-8B also lies on a favourable accuracy--parameter frontier relative to the larger general-purpose open models evaluated here and narrows the gap to the proprietary systems. Nevertheless, these gains do not eliminate the large separation from human first-pass performance, indicating that the tested data intervention is not a complete solution.}

The accuracy--parameter comparison addresses only one aspect of deployment efficiency. Work on efficient multimodal models also considers architecture, training and inference costs~\cite{jin2025efficient}; our results therefore do not establish latency or energy advantages. Memory-based navigation and exploration methods, including DORAEMON and MemoryExplorer~\cite{gu2025doraemon,wang2026memory}, offer ways to retain spatial context for sequential decisions. VideoSearcher similarly combines temporal localization, spatial focusing and external retrieval to gather evidence for video reasoning~\cite{gao2026videosearcher}. Testing whether these mechanisms improve action--goal grounding on EscherVerse is a direction for future work.

Several limitations should be noted. First, the underlying raw clips are derived from third-party online platforms, which constrains unrestricted redistribution of source videos. Second, the reported human baseline is based on retained first-pass annotation records rather than a newly designed prospective cognitive study, and we report only overall human performance in the present manuscript. Third, EscherVerse is intended to evaluate visible teleo-spatial reasoning from video evidence, not to infer private mental states or protected attributes. Future work should extend this setting to more fine-grained human evaluations, stronger temporal grounding protocols, and model architectures that explicitly represent causal dynamics, reference-frame transformations, and goal-conditioned scene structure.

\revtwo{Question-generation and filtering preferences may also leave some task types underrepresented. The new audit identifies six reference combinations without definite instances in the retained bank and shows limited within-clip overlap, despite broader type representation across the full bank. A reference drawn from the existing video categories, together with model-based coding, cannot rule out shared blind spots or classification errors. Moreover, an open-world question space has no predefined exhaustive task inventory, so a finite comparison cannot certify zero omissions. This limits claims about the completeness of coverage, but does not invalidate the observed human--model gap or the controlled comparisons on the evaluated tasks. We therefore interpret EscherVerse as a diagnostic resource for the tasks it represents and identify targeted task enrichment as a direction for future work.}

These findings have two implications for future research on vision-language and embodied AI. First, benchmark performance on static or weakly dynamic tasks should not be taken as evidence of robust spatial intelligence in real-world environments. Second, progress on teleo-spatial reasoning will likely require models that more explicitly represent causal dynamics, reference-frame transformations, and action-conditioned world structure, rather than relying on loosely coupled spatial and temporal cues. In this sense, EscherVerse serves not only as an evaluation resource, but also as a diagnostic resource for identifying where current vision-language models remain far from human-level reasoning about physical dynamics and human intent.

\rev{Safe and reliable deployment of AI in the physical world demands more than identifying objects; it also requires reasoning about physical constraints and observable human goals as scenes change. By introducing Teleo-Spatial Intelligence and the EscherVerse benchmark, we measure a substantial gap between the evaluated multimodal models and independent first-pass human performance on these tasks. Real-world, intent-aware supervision narrows but does not close this gap. Progress toward more general spatial intelligence will likely require both better training data and model designs that more explicitly represent dynamics, reference frames, and action--goal structure.}

\section*{Methods}\label{methods}

To construct a benchmark \rev{for evaluating six operational dimensions of} Teleo-Spatial Intelligence (TSI), we have developed a scalable pipeline as shown in Figure~\ref{fig:dataset}d, to generate and validate a large-scale dataset. Our methodology is specifically designed to address the key concerns of TSI, bridging the gaps between simulation and reality, static and dynamic scenes, and mere spatial description and deep intent understanding. Our pipeline consists of three integral stages: (1) multi-stage filtering for open-world video curation, (2) cognitively-inspired question-and-answer generation \rev{across these six dimensions}, and (3) multi-model cross-validation complemented by human adjudication.

\subsection*{Filter and Generation Pipeline}
To construct a video corpus rich in the complex interactions required to assess TSI, we move beyond simulated and constrained indoor environments. Real-world scenarios are essential because both Physical-Dynamic Reasoning and Intent-Driven Reasoning occur with significantly greater complexity and subtlety in open-world environments compared to controlled or curated settings.

\subsubsection*{Data Filter}
To precisely identify clips aligned with our research objectives from captions of online videos, we design a coarse-to-fine filtering pipeline:

\subsubsection*{Rule-based Filter.}
We first analyze preliminary video captions. Inspired by the scoring mechanism of Koala-36M\cite{wang2025koala}, we employ rule-based scoring and keyword matching to assign initial content scores. This approach rapidly filters out static or monotonous videos from 3,766,053 online videos, yielding over 105,000 videos.

\subsubsection*{LLM-based Filter.}
{\sloppy\looseness=-2
We employ GPT-120b-OSS\cite{agarwal2025gpt} to perform fine-grained filtering on the screened clips. Using a carefully designed red-teaming prompt, we instruct the model to score each video description, explicitly prioritizing scenes with multiple objects or multiple agents, actions, and movement, while penalizing static or ambiguous content. This process ensures that the final 11,328 selected videos exhibit high value in TSI.
\rev{These clips form the candidate pool for question generation.}
\par}

\subsubsection*{Red-Teaming QA Generation.}
{\sloppy\looseness=-2
To generate question-answer pairs that effectively probe the two core pillars of Teleo-Spatial Intelligence, we develop an automated QA generation method.
\par}

\subsubsection*{Generator Design.}
We leverage a state-of-the-art vision-language model (VLM) equipped with a specialized red-teaming expert prompt. This prompt enforces four fundamental principles to ensure question quality: prioritizing visual over linguistic challenges, grounding questions strictly in video content, avoiding shortcut questions that can be answered without viewing the video, and encouraging direct analysis rather than scene description. Additionally, we constructed a Seed Instruction Library containing high-quality, human-labeled question--answer pairs for use as seed prompts, following the seed-based generation approach of CoT-Self-Instruct~\cite{yu2025cot}. During the questioning process, seed prompts are randomly sampled from this repository and integrated into the red-teaming expert prompt. This seed repository is continuously updated with high-quality question-answer pairs generated during the question-generation process.

\subsubsection*{Scene-Oriented Questioning Strategy.}
The strategy guides the VLM to first analyze the primary driver of spatial change in a video, categorizing it as either primarily driven by human agency or by physical forces. This distinction allows us to generate questions that cover both pillars of TSI systematically:

For Human-Centric scenes, where actions are goal-oriented, the VLM is directed to generate questions that target intent-driven reasoning. For Object-Centric scenes, dominated by physical laws or mechanical processes, the VLM focuses on generating questions that probe Physical-Dynamic Reasoning.

To further refine the two pillars of our TSI concept, we have developed six dimensions of TSI, as illustrated in Figure~\ref{fig:fig1}a. All generated questions are constructed around six high-level reasoning dimensions. Collectively, they \rev{cover six operational dimensions of teleo-spatial reasoning in open-world videos}, from foundational perceptual abilities to advanced predictive and causal inference:

\begin{enumerate}
\looseness=-1
\item Action \& Intent-Driven Spatial Reasoning. Models must infer not only spatial changes but also the underlying human intentions.
\item Egocentric vs. Allocentric Reference Frames. Tests the ability to switch between and reason across different reference frames, from the camera operator's viewpoint (egocentric) to a global scene perspective (allocentric).
\sloppy\looseness=-2
\item Predictive \& Counterfactual Spatial Reasoning. Challenges models to predict future spatial states based on current trajectories and physical laws, or reason about hypothetical scenarios.
\rev{The benchmark includes both forms.}
\item Dynamic Spatial Relationships. Requires understanding of time-varying relative positions, orientations, and topological relations between objects.
\item Object Deformation \& State Transition. Evaluates understanding of non-rigid shape changes or fundamental state alterations.
\item Object Permanence \& Occlusion Tracking. Assesses the model's ability to track objects temporarily occluded or out of view continuously.
\end{enumerate}

\begin{revision}
\subsubsection*{Content categories and an item-level reasoning-demand framework.}
The six categories describe question content. For complementary analysis, we define an operational framework that distinguishes three forms of reasoning demand at the item level. Level~1 comprises temporal tracking or a current dynamic relation without prediction, reference-frame transformation or goal inference. Level~2 comprises physical prediction, counterfactual judgement or reference-frame transformation that does not depend on an actor's goal. Level~3 links an observable goal or intent to physical constraints and a present or future spatial state. We use this framework to construct and interpret the same-video paired analysis; it is not introduced as a new benchmark-wide annotation or as a fixed difficulty hierarchy.

This distinction follows accounts separating spatial visualization and orientation, intrinsic and extrinsic relations, static and dynamic processing, and spatial scale~\cite{hegarty2004dissociation,hegarty2006scales,newcombe2015typology}. EscherVerse emphasizes dynamic relations, reference-frame transformation, physical prediction and goal-conditioned spatial change. Mental rotation, large-scale navigation and static figure visualization are only partially represented.
\end{revision}

\subsubsection*{High-Quality Distractors and Chain-of-Thought Generation.}
To ensure rigorous evaluation, our prompt emphasizes the plausibility of distractors. For multiple-choice questions, incorrect options are designed to represent typical spatial cognitive errors (e.g., left-right confusion, overlooking interactions) rather than irrelevant or absurd choices. Additionally, for a substantial subset, we instruct the model to generate detailed chain-of-thought (CoT) annotations that step-by-step explain the reasoning process required to arrive at the correct answer.

\subsection*{Verification Pipeline}
Automatically generated data inevitably contains errors, ambiguities, or potential biases. \rev{To improve the quality and consistency of the retained EscherVerse items, we implement a multi-stage verification pipeline.}

Our verification pipeline employs a hierarchical, multi-stage filtering process to ensure the quality of the EscherVerse dataset. The process begins with an initial screening using Qwen2.5-VL-72B to eliminate QA pairs with fundamental flaws, such as question-answer mismatches or malformed questions. Subsequently, the filtered pairs undergo a consensus-based verification stage. We prompt Gemini-2.5-pro to perform multi-round evaluations and output its reasoning process. Inspired by self-consistency principles \cite{wang2022self}, convergence on a correct answer across multiple strong models significantly bolsters our confidence in a QA pair's quality. Conversely, disagreements flag a pair for further scrutiny. These flagged pairs are then re-evaluated by a more powerful model, Qwen3-VL-235B. The final stage involves human adjudication, where experts correct nuanced errors, remove unqualified samples, and ensure each instance stringently assesses the targeted facet of Teleo-Spatial Intelligence. To guarantee final quality, the entire candidate set is subjected to a comprehensive automated inspection by Qwen3-VL-235B and random manual audits, culminating in the final EscherVerse dataset.

{\sloppy\looseness=-1
\rev{To improve label quality among the generated candidates, the final stage involved human adjudication. A cohort of 12 trained annotators, comprising graduate researchers with backgrounds in computer vision, cognitive science, and embodied AI, participated in this verification process. This cohort was used for QA validation and quality control; it is distinct from the 11-annotator first-pass cohort analysed later for the human baseline. Annotators evaluated the generated QA pairs for physical accuracy, logical consistency of the intent-driven reasoning, and visual grounding. To quantify the reliability of this process, a random subset of 1,000 QA pairs was independently evaluated by three annotators. Fleiss' Kappa was $\kappa=0.82$, indicating high inter-annotator agreement. Samples with conflicting human judgments were reviewed and resolved by a senior meta-reviewer before inclusion. Because this procedure evaluates questions already proposed by the generation pipeline, it addresses label quality but does not by itself establish coverage of the wider question space.}
\par}

\rev{The retained construction records contain 3,569 human-review rows, of which 3,290 yield valid decisions after cleaning (Supplementary Table~10). These item-level records cover an auditable subset rather than every stage of construction. The independent scope audit described below is distinct from this verification process because it evaluates newly proposed questions rather than generated EscherVerse candidates.}

\subsection*{Final Dataset Composition}

Through the complete three-stage pipeline described above, we construct EscherVerse, comprising two primary components designed to advance the evaluation and development of models capable of Teleo-Spatial Intelligence.

The EscherVerse Benchmark. \rev{A core test set containing 8,000 question-answer pairs. This collection is designed to provide broad, but non-exhaustive, coverage of both Physical-Dynamic and Intent-Driven Reasoning and to support comparative evaluation of TSI.} The composition of the Escher-Bench is shown in Figure~\ref{fig:dataset}c.

The EscherVerse-35k. A supplementary training set containing 35,963 high-quality question-answer pairs with CoT. This dataset enables researchers to train models to learn advanced spatial reasoning, thereby understanding not only what but also how and why it is happening.

\subsection*{Independent first-pass human evaluation}
In addition to model benchmarking, we evaluated human performance using first-pass annotation records retained during benchmark construction. All benchmark instances underwent independent first-pass human annotation before consensus formation and senior adjudication. For the present analysis, we extracted only the original responses submitted before annotators had access to final labels, other annotators' answers, or model predictions. These first-pass responses therefore provide an independent human reference rather than a post hoc consensus label.

First-pass annotations were produced by 11 trained annotators following a shared protocol covering task definitions, question formats, and common failure cases in teleo-spatial reasoning. Each annotator independently viewed and answered all 8,000 benchmark questions, yielding 88,000 first-pass human responses in total, and first-pass submissions were not revised for this analysis. To avoid circular evaluation, when computing the accuracy of a given annotator we compared that annotator's response against a reference label constructed without the evaluated annotator's own answer; when exclusion of the evaluated annotator still left disagreement, a senior reviewer determined the final reference label on the basis of the raw video, the question, and the annotation guidelines. Across annotators, first-pass accuracy ranged from 84.81\% to 95.14\%, with a mean of 90.62\%. We report overall single-annotator human accuracy and the corresponding model accuracies on the same benchmark instances.

\subsection*{Model training and implementation details}
To establish a strong baseline for teleo-spatial intelligence, we developed the Escher series models (Escher-4B and Escher-8B) via Supervised Fine-Tuning (SFT). The base architectures for Escher-4B and Escher-8B are initialized from the open-source Qwen3-VL-4B and Qwen3-VL-8B base models, respectively. These models process visual inputs through a Vision Transformer (ViT) architecture fused with a causal language model backbone.

The SFT process was conducted exclusively on our proposed EscherVerse-35k dataset. We employed a Low-Rank Adaptation (LoRA)  strategy to adapt the spatial reasoning capabilities. To preserve spatiotemporal continuity, video inputs were uniformly sampled at 16 frames and processed dynamically.

Unless otherwise stated, all reported Escher models were trained on the EscherVerse-35k instruction-tuning set using the same frame-sampling protocol and answer format adopted at evaluation time.

\subsection*{Evaluation setup and scoring}
We measured accuracy by directly comparing generated answers with ground truth without external evaluator models. To ensure deterministic and comparable results, all models were evaluated using a unified zero-shot protocol with a temperature of 0 and a maximum output length of 8,192 tokens. For video inputs, we applied uniform temporal sampling of 16 frames per video. Models were instructed to enclose final answers within \texttt{<answer>} tags for automated parsing. Scoring logic varied by question type: single/multiple-choice questions used hard matching of option letters; True/False questions used keyword matching; fill-in-the-blank questions awarded full credit for exact matches and partial credit based on text similarity thresholds. Human-model comparisons were conducted on the same benchmark instances with retained first-pass human responses.

\subsection*{Statistical analyses}
\begin{revision}
\subsubsection*{Correlation and component analysis.}
{\looseness=-1
We formed a $27\times6$ matrix from the category accuracies in Table~\ref{tab:spatial_reasoning_benchmark} and computed all 15 Spearman correlations with model-resampling intervals. After standardizing each category across models, we applied principal-component analysis. Parallel analysis retained a component when its eigenvalue exceeded the corresponding 95th-percentile eigenvalue from random matrices of the same dimensions. Leave-one-model-out analysis assessed stability.\par}

\subsubsection*{Perception controls and same-video pairs.}
The perception panel contains 200 distinct benchmark videos. Each control tests entity presence or identity, action or event recognition, or a simple relation visible in the same 16-frame input, without reproducing the original higher-level inference. We report paired differences and exact two-sided McNemar tests. The same-video panel contains 50 videos: 17 pairs compare dynamics with intent, 17 compare camera-centred with actor-goal reasoning, and 16 compare tracking with prediction. We report source-video-clustered intervals and a two-sided sign-flip test over within-video differences.

\subsubsection*{Equal-budget SFT ablation.}
{\looseness=-1
The Intent-focused, Dynamic-only and Random-mixture conditions contain 1,409 examples each. We removed source-video overlap with the benchmark and matched the subsets on question format and normalized gold answer. All conditions use Qwen3-VL-4B-Instruct with identical LoRA capacity, optimization steps, frame policy and two seeds. Evaluation uses disjoint 486-item Intent and Dynamic panels. A 10,000-replicate bootstrap tests this contrast.\par}

\subsubsection*{Behavioral error analysis.}
{\looseness=-1
We selected three models and sampled 100 errors per model, proportionally stratified by the six categories. A frozen codebook defines eight mutually exclusive primary labels: benchmark ambiguity, protocol/parsing, perception/recognition, spatiotemporal grounding, perspective/reference frame, action--goal binding, physical prediction/counterfactual and other unresolved. Two independent raters blinded to model identity assigned labels, and a third rater adjudicated disagreements. Inverse sampling-probability weights recover each source error pool's category composition, with a Kish adjustment for unequal weights. The intervals are descriptive because 300 model--item errors represent 169 unique benchmark items.\par}

\subsubsection*{Independent scope audit.}
We sampled 20 videos per category, for 120 distinct videos. Two independent authoring runs received 64 uniformly sampled full-span frames but no EscherVerse question, answer or category label, and each proposed three answerable questions per video. After evidence and duplication checks, 666 questions remained. Two independent scope raters classified each question as directly covered, a supporting operation or outside scope. \revtwo{The primary taxonomy-alignment estimate requires both raters to mark a question as directly covered or supporting; it does not measure representation in the retained question bank.} We report agreement statistics and 10,000-replicate bootstrap intervals clustered by source video.
\end{revision}

\subsection*{\revtwo{Audit of retained task types}}
\revtwo{Human annotators wrote 200 reference questions from 60 clips drawn from distinct source videos, with 10 clips per benchmark category and three or four questions per clip. They received the overall task scope and anonymized sampled frames, but no original questions, answers or clip-level sampling labels. They viewed 64 frames for 55 clips and 20 frames for the other five, rather than continuous full videos. Thirty clips had been used in the earlier analysis. We sampled 200 original questions, comprising 100 from videos used for the reference questions and 100 from other videos, and also examined all 8,000 original records.}

\revtwo{A codebook developed from the 200 reference and 200 sampled original questions distinguishes 17 content categories and 13 required operations, with subtypes where appropriate. Each content--operation combination must be supported by the task requested, and a question can receive multiple labels. For full-bank coding, \texttt{gpt-6-sol} received question text without answers, videos or the original six-category labels. The bank contains 6,283 unique clip--question groups; identical records share a coding result, mapped back to the original records for frequency reporting. Only definite labels contribute to the counts; tentative labels are retained separately. This initial model-based classification has not undergone complete human annotation review and does not verify answers.}

\revtwo{We repeated the comparison after excluding the 193 flagged original records, after excluding flagged questions from both sets, and after restricting the reference to either 64-frame inputs or the 30 clips not used in the earlier analysis. For within-clip overlap, we compared each of the 186 unflagged reference questions with all retained original questions for the same clip. Supplementary Note~8 details the counting rules and limitations. Because the codebook was developed from the study samples, these comparisons are descriptive rather than independent validation of a prespecified taxonomy.}

\subsubsection*{Ethics and bias considerations.}
EscherVerse is constructed from short clips derived from publicly accessible online videos depicting real-world human and object interactions. The dataset is intended for research on multimodal reasoning, not for identity recognition, person re-identification, or profiling of individuals. Because the benchmark includes human-centred scenes and questions about goal-directed behaviour, annotators were instructed to ground answers only in observable spatiotemporal evidence, physical context, and directly supported action goals, and to reject samples that required speculative inference about private mental states, sensitive personal attributes, or ambiguous intentions. During verification, samples judged ethically sensitive, weakly grounded, or insufficiently supported by the video evidence were removed. We discuss the benchmark as a measure of visible teleo-spatial reasoning rather than a direct readout of human psychology, and we caution against deploying it to infer protected attributes or internal states from real-world videos.

\section*{Data Availability}
The EscherVerse benchmark annotations (\texttt{Escher-Bench.json}), instruction-tuning data (\texttt{Escher-sft.jsonl}), preference-training subset (\texttt{Escher-GRPO-Subset.jsonl}) and video metadata (\texttt{video\_list.json}) are publicly available at \url{https://huggingface.co/datasets/Gradygu3u/EscherVerse-Data}. The metadata provide source YouTube URLs and precise start and end times for 11,328 clips. The project repository at \url{https://github.com/Grady10086/EscherVerse} provides instructions and a clip-reconstruction script that can use retained source files or retrieve source videos that remain publicly accessible. The released repositories do not host the video files themselves, and reconstruction depends on source availability and applicable platform terms. Because the source videos are third-party content, access to retained clips is controlled and subject to availability and source-platform terms. Qualified academic researchers may request access for non-commercial research from the corresponding author; editors and referees can request access to the materials needed to evaluate the manuscript. If a video cannot be accessed through the public sources, researchers may contact the corresponding author to request access to a retained copy. We will respond to such requests within 10 working days and, upon approval, provide downloadable copies with no additional time limit on their use for the approved research purposes. Source data are provided with this paper. The Source Data file contains the numerical data supporting the figures and tables, with an index linking each display item to its data and a description of the available records and aggregation methods.

\section*{Code Availability}
The project repository at \url{https://github.com/Grady10086/EscherVerse} provides the maintained benchmark evaluator and common scorer, video-reconstruction tooling, code and frozen artifacts for the revision analyses, and the SFT/LoRA path used for the content-specific ablation. The code and reproducibility resources are archived on Zenodo (version v1.0.0; DOI: \url{https://doi.org/10.5281/zenodo.23877838})\cite{gu2026escherversecode}. A model registry distinguishes results that support fresh inference from those available only as configuration references or reported aggregates. The human-baseline release provides an aggregation script and a manuscript-level aggregate validation record; it does not include item-level judgments.

\section*{Funding Statement}
\revtwo{This work was supported by the Shanghai Committee of Science and Technology (Grant Nos. 25511103300, 25511104302, 25511102700), the National Natural Science Foundation of China (Grant Nos. 62302167, U23A20343, W2521174, 62672185, 62476090, 62502159, 62676147), and the Young Elite Scientists Sponsorship Program by CAST (Grant No. YESS20240780).}

\section*{Author Contributions Statement}
T.G. conceived the study, designed and developed the benchmark pipeline, conducted the model evaluations, analysed the results, and drafted the manuscript. J.G., Z.Z., Y.X., and L.M. contributed to data collection, curation, annotation, and benchmark verification. X.T. and A.V.V. supervised the work, guided the study design and interpretation of the results, and revised the manuscript. All authors reviewed and approved the final manuscript.

\section*{Ethics declaration}
Not applicable.

\section*{Competing Interests Statement}
The authors declare no competing interests.

\clearpage
\section*{Tables}
\begin{table*}[!ht]
\centering
\small 
\caption{\textbf{Comprehensive evaluation of Vision-Language Models on the spatial reasoning benchmark.} Accuracy (\%) is reported for all models. Independent human performance is reported as overall first-pass accuracy only, because category-level human breakdowns are not part of this table. Source data are provided as a Source Data file.}
\resizebox{\textwidth}{!}{
\setlength\tabcolsep{4pt}
\renewcommand\arraystretch{1.3}
\begin{tabular}{l || cccccc | c | cc | cccc}
\toprule

\normalsize 
\textbf{Model} & \multicolumn{6}{c|}{\textbf{Spatial Reasoning Abilities}} & \textbf{Overall} & \multicolumn{2}{c|}{\textbf{Centricity}} & \multicolumn{4}{c}{\textbf{Question Types}} \\
\cmidrule(lr){2-7} \cmidrule(lr){8-8} \cmidrule(lr){9-10} \cmidrule(lr){11-14}

 & \shortstack{Obj. Perm. \\ \& Occl.} & \shortstack{Dynamic \\ Spatial} & \shortstack{Action \& \\ Intent} & \shortstack{Predictive \\ \& Count.} & \shortstack{Obj. Def. \\ \& State} & \shortstack{Ego. vs. \\ Allo.} & \shortstack{Overall \\ Acc.} & \shortstack{Human- \\ Centric} & \shortstack{Object- \\ Centric} & \shortstack{Fill-in \\ Blank} & \shortstack{Multi- \\ Select} & \shortstack{Single \\ Choice} & \shortstack{True/ \\ False} \\
\midrule
\midrule

\multicolumn{14}{l}{\textit{Open-Source Models}} \\
\midrule
Qwen3-VL-2B-Instruct	& 39.87	& 38.24	& 43.96	& 38.55	& 39.34	& 37.90	& 38.90	& 39.88	& 35.74	& 30.72	& 7.80	& 34.03	& 50.35 \\

Qwen2.5-VL-3B-Instruct	& 39.13	& 38.98	& 47.13	& 41.89	& 44.55	& 40.15	& 40.61	& 41.12	& 38.96	& 36.41	& 11.36	& 35.46	& 50.99 \\
Qwen2.5-VL-7B-Instruct	& 43.37	& 37.31	& 46.83	& 40.03	& 39.34	& 34.90	& 38.66	& 38.49	& 39.22	& 27.96	& 13.36	& 33.07	& 50.74 \\

Qwen3-VL-4B-Instruct 	& 49.17	& 44.47	& 51.66	& 44.98	& 54.03	& 42.96	& 45.60	& 46.06	& 44.12	& 47.51	& 34.08	& 38.44	& 53.93 \\
Qwen3-VL-4B-Thinking	& 47.42	& 44.39	& 51.51	& 44.73	& 56.87	& 47.02	& 46.58	& 48.07	& 41.75	& 51.71	& 35.19	& 40.83	& 52.57 \\

Qwen3-VL-8B-Instruct	& 49.82	& 44.51	& 57.40	& 44.81	& 60.66	& 43.08	& 46.34	& 46.91	& 44.49	& 48.29	& 35.86	& 38.80	& 54.90 \\
Qwen3-VL-8B-Thinking	& 44.48	& 45.64	& 52.42	& 42.26	& 60.19	& 47.49	& 46.45	& 47.91	& 41.75	& 52.49	& 41.43	& 40.44	& 51.59 \\

Qwen3-VL-30B-A3B-Instruct	& 50.55	& 44.99	& 56.50	& 44.32	& 59.72	& 45.22	& 47.06	& 47.52	& 45.60	& 47.73	& 37.64	& 40.08	& 55.27 \\

Qwen3-VL-32B-Thinking	& 49.26	& 49.80	& 58.76	& 49.34	& 60.19	& 46.04	& 49.58	& 49.70	& 49.18	& 57.68	& 48.55	& 43.30	& 53.87 \\
LLaVA-OneVision-1.5-4B-Instruct	& 42.17	& 42.78	& 51.21	& 41.60	& 52.13	& 43.73	& 43.75	& 44.37	& 41.75	& 45.08	& 36.30	& 39.49	& 48.74 \\

LLaVA-OneVision-1.5-8B-Instruct	& 39.59	& 44.15	& 50.60	& 42.42	& 51.18	& 45.18	& 44.30	& 45.17	& 41.49	& 43.20	& 36.75	& 40.05	& 49.95 \\
MiniCPM-V-4	& 43.92	& 40.13	& 47.73	& 38.55	& 35.55	& 40.39	& 40.99	& 41.18	& 40.38	& 35.58	& 23.61	& 37.67	& 48.22 \\

InternVL3-8B & 43.00 & 42.94 & 52.27 & 41.43 & 42.18 & 43.25 & 43.56 & 44.55 & 40.38 & 43.20 & 16.48 & 39.90 & 51.08 \\
ERNIE-4.5-VL-28B-A3B-Thinking	& 41.34	& 40.17	& 51.66	& 40.28	& 57.35	& 41.22	& 42.06	& 43.11	& 41.08	& 49.72	& 35.19	& 37.85	& 49.01 \\

LLaVA-NeXT-Video-7B	& 6.26	& 10.28	& 8.91	& 16.05	& 2.84	& 13.66	& 11.30	& 10.63	& 13.44	& 1.10	& 6.46	& 25.80	& 0.00 \\
Spatial-MLLM	& 10.04	& 14.44	& 26.74	& 22.80	& 9.48	& 13.83	& 15.81	& 15.86	& 15.66	& 17.24	& 1.34	& 32.90	& 0.00 \\

ViCA2-7B	& 9.67	& 15.24	& 27.79	& 23.97	& 14.69	& 13.92	& 16.45	& 15.96	& 18.03	& 17.57	& 2.00	& 34.21	& 0.03 \\
\midrule

\multicolumn{14}{l}{\textit{Open-Source Models (Supervised Fine-Tuning)}} \\
\midrule

ISI-8B-Instruct	& 46.59	& 43.01	& 56.50	& 43.46	& 50.24	& 43.64	& 45.06	& 46.06	& 41.86	& 50.61	& 26.50	& 38.92	& 52.32 \\

Escher-7B-Instruct	& 47.61	& 48.88	& 62.08	& 48.89	& 53.55	& 45.31	& 48.89	& 49.17	& 47.97	& 51.05	& 41.20	& 46.88	& 51.38 \\

Escher-4B-Instruct	& 49.72	& 48.59	& 61.63	& 47.33	& 52.61	& 47.71	& 49.49	& 49.73	& 48.71	& 51.82	& 41.87	& 47.00	& 52.41 \\

Escher-4B-Thinking	& 48.07	& 47.55	& 61.33	& 47.32	& 53.08	& 46.38	& 48.54	& 48.86	& 47.49	& 49.72	& 43.65	& 45.21	& 52.26 \\

Escher-8B-Instruct	& 46.96	& 49.80	& 63.60	& 48.31	& 56.87	& 47.50	& 49.85	& 50.88	& 46.55	& 52.60	& 48.11	& 46.08	& 53.17 \\

Escher-8B-Thinking	& 46.04	& 48.67	& 62.84	& 48.64	& 60.19	& 45.22	& 48.79	& 49.12	& 47.71	& 50.28	& 44.99	& 46.08	& 51.66 \\
\midrule
\midrule

\multicolumn{14}{l}{\textit{Closed-Source Models (API-based)}} \\
\midrule
qwen3-vl-plus	& 44.38	& 43.21	& 55.52	& 44.61	& 57.82	& 42.38	& 44.75	& 45.83	& 41.28	& 44.86	& 44.77	& 38.63	& 50.94 \\

GPT-5	& 52.24	& 53.28	& 64.66	& 50.41	& 67.99	& 50.58	& 53.24	& 54.28	& 49.88	& 60.01	& 55.43	& 47.96	& 56.49 \\
Gemini-2.5-flash	& 53.13	& 51.71	& 65.01	& 49.22	& 65.88	& 48.91	& 52.19	& 52.93	& 49.66	& 56.02	& 48.55	& 45.51	& 58.42 \\

Gemini-2.5-pro	& 56.17	& 57.29	& 69.53	& 54.20	& 73.11	& 54.39	& 57.26	& 58.37	& 53.64	& 64.53	& 59.60	& 51.57	& 60.74 \\
\midrule
Human first-pass (overall only, n=11) & \multicolumn{6}{c|}{--} & 90.62 (84.81--95.14) & \multicolumn{2}{c|}{--} & \multicolumn{4}{c}{--} \\
\bottomrule
\end{tabular}%
} 
\label{tab:spatial_reasoning_benchmark}
\end{table*}
\clearpage
\begin{table*}[!ht]
\centering
\small
\caption{Equal-budget SFT ablation. Parentheses report the standard deviation across two training seeds rather than confidence intervals. Source data are provided as a Source Data file.}
\begin{tabular}{lrrrr}
\toprule
Condition & Intent & $\Delta$ Base & Dynamic & $\Delta$ Base \\
\midrule
Base & 50.21\% & -- & 37.65\% & -- \\
Intent-focused & 57.51\% (3.35) & +7.30 & 39.09\% (0.00) & +1.44 \\
Dynamic-only & 54.53\% (1.16) & +4.32 & 43.62\% (1.16) & +5.97 \\
Random mixture & 52.16\% (1.02) & +1.95 & 41.77\% (1.45) & +4.12 \\
\bottomrule
\end{tabular}
\label{tab:sft_ablation}
\end{table*}
\clearpage
\begin{table*}[!ht]
\centering
\small
\caption{Adjudicated primary error distribution. Weighted intervals use a Kish effective sample size and are descriptive. Source data are provided as a Source Data file.}
\begin{tabular}{lrrr}
\toprule
Primary label & $n$ & Weighted & 95\% interval \\
\midrule
Spatiotemporal grounding & 86 & 28.8\% & 24.0--34.2 \\
Perspective/reference frame & 72 & 23.7\% & 19.3--28.9 \\
Physical prediction & 43 & 14.2\% & 10.7--18.6 \\
Perception/recognition & 36 & 12.1\% & 8.9--16.3 \\
Benchmark ambiguity & 29 & 9.6\% & 6.8--13.5 \\
Action--goal binding & 22 & 7.4\% & 5.0--11.0 \\
Protocol/parsing & 12 & 4.1\% & 2.4--7.0 \\
Other unresolved & 0 & 0.0\% & 0.0--1.3 \\
\bottomrule
\end{tabular}
\label{tab:error_distribution}
\end{table*}
\end{document}